\pdfoutput=1

\documentclass[11pt]{article}

\usepackage{acl}

\usepackage{times}
\usepackage{latexsym}

\usepackage[T1]{fontenc}

\usepackage[utf8]{inputenc}

\usepackage{microtype}

\usepackage{inconsolata}

\usepackage{multirow}
\usepackage{latexsym}
\usepackage{booktabs}
\usepackage{courier}
\usepackage{amsmath}
\usepackage{amsfonts} 
\usepackage{subfig}
\usepackage{placeins}
\usepackage{tikz}
\usepackage{hyperref}
\usepackage{algorithm}
\usepackage{algpseudocode}

\title{Multi-Operational Mathematical Derivations in Latent Space}

\author{Marco Valentino$^1$, Jordan Meadows$^2$, Lan Zhang$^2$, Andr\'e Freitas$^{1,2,3}$ \\ $^1$Idiap Research Institute, Switzerland \\ 
$^{2}$ Department of Computer Science, University of Manchester, United Kingdom\\
$^{3}$ Cancer Biomarker Centre, CRUK Manchester Institute, United Kingdom\\ 
\texttt{\{marco.valentino, andre.freitas\}@idiap.ch} \\
\texttt{\{jordan.meadows,  lan.zhang-6\}@manchester.ac.uk}}

\begin{document}

\maketitle

\begin{abstract}
This paper investigates the possibility of approximating multiple mathematical operations in latent space for expression derivation. To this end, we introduce different multi-operational representation paradigms, modelling mathematical operations as explicit geometric transformations. 
By leveraging a symbolic engine, we construct a large-scale dataset comprising 1.7M derivation steps stemming from 61K premises and 6 operators, analysing the properties of each paradigm when instantiated with state-of-the-art neural encoders.
Specifically, we investigate how different encoding mechanisms can approximate expression manipulation in latent space, exploring the trade-off between learning different operators and specialising within single operations, as well as the ability to support multi-step derivations and out-of-distribution generalisation.
Our empirical analysis reveals that the multi-operational paradigm is crucial for disentangling different operators, while discriminating the conclusions for a single operation is achievable in the original expression encoder. Moreover, we show that architectural choices can heavily affect the training dynamics, structural organisation, and generalisation of the latent space, resulting in significant variations across paradigms and classes of encoders\footnote{Code \& data available at: \url{https://github.com/neuro-symbolic-ai/latent_mathematical_reasoning}}.
\end{abstract}

\section{Introduction}

\begin{figure}[htp]
    \centering
    \includegraphics[width=0.85\columnwidth]{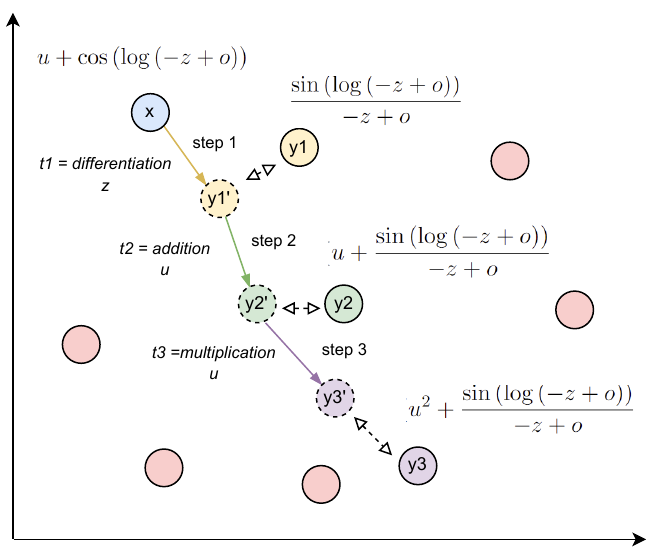}
    \caption{
    \emph{Can neural encoders learn to approximate multiple mathematical operators in latent space?} Given a premise $x$, we investigate the problem of applying a sequence of latent operations ($t_1,\ldots,t_n$) to derive valid mathematical expressions ($y_1,\ldots,y_n$).
    }
\label{fig:introduction}
\end{figure}

To what extent are neural networks capable of mathematical reasoning? This question has led many researchers to propose various methods to train and test neural models on different math-related tasks, such as math word problems, theorem proving, and premise selection \cite{lu-etal-2023-survey,meadows2023introduction,mishra-etal-2022-lila,ferreira-etal-2022-integer,ferreira2020premise,welleck2021naturalproofs,valentino-etal-2022-textgraphs,mishra2022numglue,petersen-etal-2023-neural}. These methods aim to investigate how neural architectures learn and generalise mathematical concepts and symbolic rules, and how they cope with characteristic challenges of mathematical inference, such as abstraction, compositionality, and systematicity \cite{welleck2022symbolic,mishra-etal-2022-lila}.

In general, a key challenge in neural mathematical reasoning is to represent expressions and formulae into a latent space to enable the application of multiple operations in specific orders under contextual constraints. Existing methods, however, typically focus on single-operational inference -- i.e., optimising a latent space to approximate a specific mathematical operation \cite{lee2019mathematical,lample2019deep,welleck2022symbolic}. Encoding multiple operations in the same latent space, therefore, remains an unexplored challenge that will likely require the development of novel mechanisms and representational paradigms.

To investigate this problem, this paper focuses on \emph{equational reasoning}, intended as \emph{the derivation of expressions from premises via the sequential application of specialised mathematical operations} (i.e., addition, subtraction, multiplication, division, integration, differentiation). As derivations represent the workhorse of applied mathematical reasoning (including derivations in physics and engineering), projecting expressions and operators into a well-organised geometric space can unveil a myriad of applications, unlocking the approximation of mathematical solutions that are multiple steps apart within the embedding space via distance metrics and vector operations.

Specifically, this paper posits the following overarching research questions: \emph{RQ1:``How can different representational paradigms and encoding mechanisms support expression derivation in latent space?''; 
RQ2:``What is the representational trade-off between generalising across different mathematical operations and specialising within single operations?''; 
RQ3:``To what extent can different encoding mechanisms enable multi-step derivations through the sequential application and functional composition of latent operators?''; RQ4:``To what extent can different encoding mechanisms support out-of-distribution generalisation?'' }

To answer these questions, we investigate joint-embedding predictive architectures \cite{lecun2022path} by introducing different multi-operational representation paradigms (i.e., \emph{projection} and \emph{translation}) to model mathematical operations as explicit geometric transformations within the latent space.
\begin{figure*}[htp]
    \centering
    \includegraphics[width=0.9\textwidth]{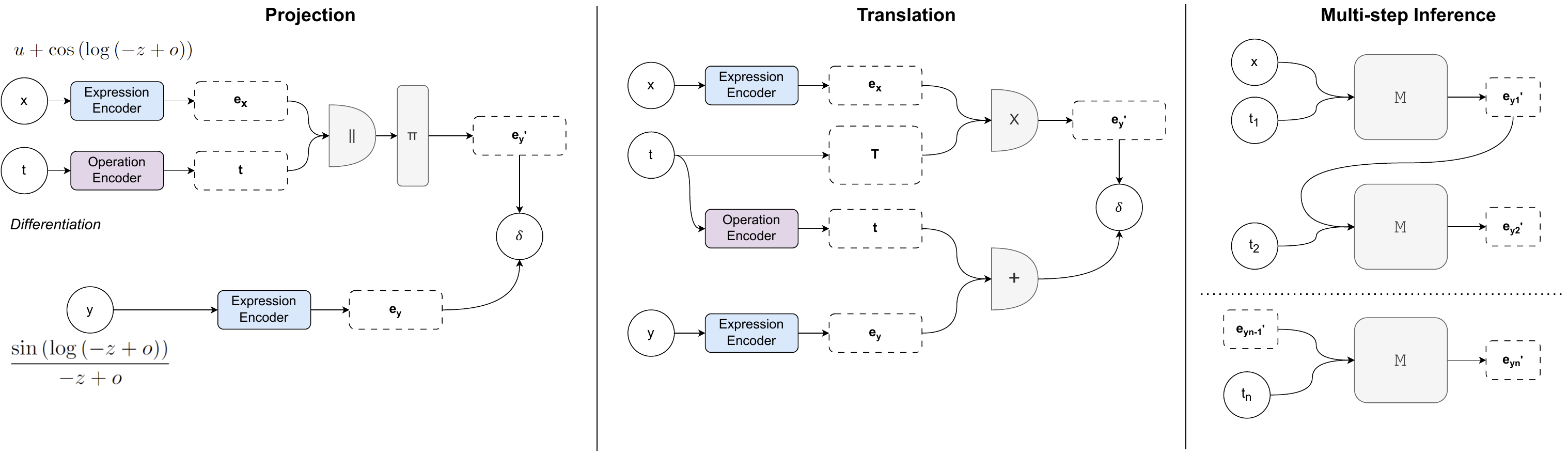}
    \caption{
    Overview of the proposed joint-embedding predictive architectures for latent multi-operational derivation (left). Schematic workflow for multi-step inference and latent propagation of mathematical operations (right).
    }
\label{fig:architectures}
\end{figure*}
Moreover, by leveraging a symbolic engine \cite{meurer2017sympy}, we build a large-scale dataset containing 1.7M derivation steps which span diverse mathematical expressions and operations. To understand the impact of different encoding schemes on equational reasoning, we instantiate the proposed architectures with state-of-the-art neural encoders, including Graph Neural Networks (GNNs) \cite{hamilton2017inductive,kipf2016semi}, Convolutional Neural Networks (CNNs) \cite{li2021survey,kim2014convolutional}, Recurrent Neural Networks (RNNs) \cite{yu2019review,hochreiter1996lstm}, and Transformers \cite{vaswani2017attention}, analysing the properties of the latent spaces and the ability to support multi-step derivations and generalisation.

Our empirical evaluation reveals that the multi-operational paradigm is crucial for disentangling different mathematical operators (i.e., \emph{cross-operational inference}), while the discrimination of the conclusions for a single operation (i.e., \emph{intra-operational inference}) is achievable in the original expression encoder. Moreover, we show that architectural choices can heavily affect the training dynamics and the structural organisation of the latent space, resulting in significant variations across paradigms and classes of encoders.

Overall, we conclude that the translation paradigm can result in a more fine-grained and smoother optimisation of the latent space, which better supports cross-operational inference and enables a more balanced integration. Regarding the encoders, we found that sequential models achieve more robust performance when tested on multi-step derivations, while graph-based encoders, on the contrary, exhibit better generalisation to out-of-distribution examples.   


\section{Multi-Operational Derivations}
\label{sec:multi_operational_inference}
Given a premise $x$ -- i.e., a mathematical expression including variables and constants, and a set of operations $T = \{t_1, t_2, \ldots t_n\}$ -- e.g., addition, multiplication, differentiation, etc., we investigate the extent to which a neural encoder can approximate a mathematical function $f(x,t_i;V) = Y_{t_i}$ that takes the premise $x$ and any operation $t_i \in T$ as inputs, and produces the set of valid expressions $Y_{t_i} = \{y_1, y_2, \ldots, y_m\}$ derivable from $x$ via $t_i$ given $v \in V$, where $V$ is a predefined set of operands such that $y_j = t_i(x, v_j)$. In this work, we focus on \emph{atomic operations} in which $V$ includes symbols representing variables. 

For example, consider the following premise $x$: $$u + \cos{(\log{(- z + o)})}$$ 
If the set of operands is $V =\{z,u\}$, and the operation $t_i$ is \emph{addition}, then the application of $t_i$ to $x$ should result in the set of expressions:

$$Y_{add} = \{z + u + \cos{(\log{(- z + o)})}, \ldots, $$$$ 2 u + \cos{(\log{(- z + o)})}\}$$

Instead, if $t_i$ is \emph{differentiation}, then $f(x,t_i;V)$ should result in a different set: $$Y_{diff} = \{ \frac{\sin{(\log{(- z + o)})}}{- z + o}, \ldots, 1\}$$
Notably, the recursive application of $f$ to any of the expressions in $Y_{t_i}$ can generate a new set of conclusions derivable from $x$ in multiple steps.

Here, the constraint we are interested in, is that $Y_{add}$ and $Y_{diff}$ should be derived via a \emph{single expression encoder} that maps expressions into a vector space and, at the same time, enables a multi-step propagation of latent operations.

\subsection{Architectures}
\label{sec:jepa}
To model latent mathematical operations, we investigate the use of joint-embedding predictive architectures \cite{lecun2022path}. In particular, we introduce two multi-operational paradigms based on \emph{projection} and \emph{translation} to learn the representation of expressions and mathematical operators and model an atomic derivation step as an explicit \emph{geometric transformation}. 
Figure \ref{fig:architectures} shows a schematic representation of the architectures.

In general, both projection and translation employ an \emph{expression encoder} to map the premise $x$ and a plausible conclusion $y$ into vectors, along with an \emph{operation encoder} that acts as a \emph{latent prompt} \textbf{t} to discriminate between operators. The goal is then to predict the embedding of a valid conclusion $\textbf{e}_y$ by applying a transformation to the premise embedding $\textbf{e}_x$ conditioned on \textbf{t}. Therefore, the two paradigms mainly differ in how expression and operation embeddings are combined to approximate the target results. This setup enables \emph{multi-step inference} since the predicted embedding $\textbf{e}_y'$ can be recursively interpreted as a premise representation for the next iteration (Figure \ref{fig:architectures}, right). 

\paragraph{Projection.} The most intuitive solution to model latent mathematical operations is to employ a \emph{projection layer} \cite{lee2019mathematical}. In this case, the premise $x$ and the operator $t$ are first embedded using the respective encoders, which are then fed to a dense predictive layer $\pi$ to approximate the target conclusion $\textbf{e}_y$. The overall objective function can then be formalised as follows:

\begin{equation}
    \label{eq:projection}
    \begin{aligned}
        \phi(x,t,y) = -\delta(\pi(\textbf{t}\|\textbf{e}_{x}), \textbf{e}_y)^2
    \end{aligned}
\end{equation}
Where $\delta$ is a distance function, and $\pi$ represents the dense projection applied to the concatenation $\|$ of $\textbf{t}$ and $\textbf{e}_x$. While many options are available, we implement $\pi$ using a linear layer to better investigate the representation power of the underlying expression encoder.

\paragraph{Translation.} Inspired by research on multi-relational graph embeddings \cite{bordes2013translating,balazevic2019multi,valentino2023multi}, we frame mathematical inference as a \emph{multi-relational representation learning} problem. In particular, it is possible to draw a direct analogy between entities and relations in a knowledge graph and mathematical operations. Within the scope of the task, as defined in Section 2, the application of a general operation can be interpreted as a relational triple $< x,t,y >$, in which a premise expression $x$ corresponds to the subject entity, a conclusion $y$ corresponds to the object entity, and the specific operation type $t$ represents the semantic relation between entities.
Following this intuition, we formalise the learning problem via a translational objective:

\begin{equation}
    \label{eq:translation}
    \begin{aligned}
        \phi(x,t,y) = -\delta(\textbf{T}\textbf{e}_x, \textbf{e}_y + \textbf{t})^2
    \end{aligned}
\end{equation}
Where $\delta$ is a distance function, $\textbf{e}_x$, $\textbf{e}_y$, \textbf{t}, are the embeddings of premise expression, conclusion and operation, and \textbf{T} is a diagonal operation matrix.

\subsection{Data Generation}
\label{sec:synthetic_data}
We generate synthetic data to support the exploration of the above architectures, inspired by a recent approach that relies on a symbolic engine to generate equational reasoning examples~\cite{meadows2023symbolic}. In particular, we use SymPy \cite{meurer2017sympy} to construct a dataset containing expressions in both LaTeX and SymPy surface forms. Here, premises and variables are input to 6 operations to generate further expressions, presently focusing on differentiation, integration, addition, subtraction, multiplication, and division. Concrete examples of entries in the dataset are reported in the Appendix.

\paragraph{Premises.} To generate a premise expression, a set of symbols is first sampled from a vocabulary. Subsequently, an initial operator is applied to symbols to generate an expression via the SymPy engine. To generate more complex expressions, this process is repeated iteratively for a fixed number of steps. 
This process is formalised in Algorithm~\ref{alg:randomisation} (see Appendix). The final dataset includes 61K premises each containing between 2 to 5 variables. 

\paragraph{Applying Operations to Premises.} For a given premise, a set of operand variables (denoted by $V$ in Section~\ref{sec:multi_operational_inference}) are sampled from the vocabulary and added to the set of symbols that comprise the premise. All valid combinations of premise and operands are then input to each operator (via SymPy) to generate conclusions derivable via atomic derivation steps. The resulting dataset contains a total of 1.7M of such atomic steps. This data is used to train and evaluate models on single-step inference before testing generalisation capabilities to multiple steps.

\paragraph{Multi-Step Derivations.} To test the models' ability to derive expressions obtained after the sequential application of operations, we randomly sample 5K premises from the single-step dataset described above and iteratively apply up to 6 operations to each premise using a randomly sampled variable operand from the vocabulary for each step.
We adopt this methodology to generate a total of 2.7K multi-step examples.

\subsection{Expression Encoders}
Thanks to their generality, the multi-operational architectures can be instantiated with different classes of expression encoders. In particular, we experiment with both \emph{graph-based} and \emph{sequential} models, exploring embeddings with different dimensions (i.e., 300, 512, and 768). The graph-based encoders are trained on operation trees extracted from the SymPy representation, while the sequential models are trained on LaTeX expressions. We adopted the following expression encoders in our experiments:

\paragraph{Graph Neural Networks (GNNs).}  GNNs have been adopted for mathematical inference thanks to their ability to capture explicit structural information \cite{lee2019mathematical}. Here, we consider different classes of GNNs to experiment with models that can derive representations from operation trees. Specifically, we employ a 6-layer GraphSage\footnote{\url{https://pytorch-geometric.readthedocs.io/en/latest/generated/torch_geometric.nn.models.GraphSAGE.html}}\cite{hamilton2017inductive} and Graph Convolutional Network (GCN)\footnote{\url{https://pytorch-geometric.readthedocs.io/en/latest/generated/torch_geometric.nn.models.GCN.html}} \cite{kipf2016semi} to investigate transductive and non-transductive methods. To build the operation trees, we directly parse the SymPy representation described in Appendix \ref{sec:details_generation}.

\paragraph{Convolutional Neural Networks (CNNs).} CNNs represent an effective class of models for mathematical representation learning thanks to their translation invariance property that can help localise recurring symbolic patterns within expressions \cite{petersen-etal-2023-neural}. Here, we employ a 1D CNN architecture typically used for text classification tasks \cite{kim2014convolutional}, with three filter sizes 3, 4, and 5, each with 100 filters.

\paragraph{Recurrent Neural Networks (RNNs).} Due to the sequential nature of mathematical expressions, we experiment with RNNs that have been successful in modelling long-range dependencies for sentence representation \cite{yu2019review,hochreiter1996lstm}. In particular, we employ a Long-Short Term Memory (LSTM) network with 2 layers.

\paragraph{Transformers.} Finally, we experiment with a Transformer encoder with 6 and 8 attention heads and 6 layers, using a configuration similar to the one proposed by \citet{vaswani2017attention}\footnote{\url{https://pytorch.org/docs/stable/generated/torch.nn.TransformerEncoder.html}}. Differently from other models, Transformers use the attention mechanism to capture implicit relations between tokens, allowing, at the same time, experiments with a larger number of trainable parameters.


\begin{table*}[t]
    \tiny
    \centering
    \begin{tabular}{l|ccc|ccc|c}
        \toprule
         \textbf{}  & \textbf{MAP} & \textbf{Hit@1} & \textbf{Hit@3} & \textbf{MAP} & \textbf{Hit@1} & \textbf{Hit@3} & \textbf{Avg. MAP}\\ 
       \midrule
       \textbf{Projection (One-hot)} & & \textbf{Cross-op.} & & & \textbf{Intra-op.} &\\
       \midrule
       GCN & 74.07 & 78.35 & 88.88 & 93.34 & 97.81 & 99.01 & 83.70 \\
       GraphSAGE & \textbf{83.89} & \textbf{88.43} & \textbf{96.70} & 93.00 & 97.45 & 98.71 & \textbf{88.44} \\
       \midrule
       CNN& 69.61 & 76.98 & 95.20 & 92.43 & 97.18 & 98.63 &  81.02\\
       LSTM & 71.40 & 73.50 & 90.08 & \textbf{93.21} & \textbf{98.01} & \textbf{99.35} & 82.30\\
       Transformer& 49.35 & 46.30 & 63.00 & 91.66 & 96.65 & 98.38 & 70.50\\
       \midrule
       \textbf{Projection (Dense)} & & & & & & \\
        \midrule
       GCN & 78.25 & 82.50 & 92.81 & 93.43 & 97.91 & 99.08 & 85.84\\
       GraphSAGE & 81.05 & 83.91 & 94.38 & 93.18 & 97.81 & 98.93 & 87.11\\
       \midrule
       CNN& \textbf{82.57} & \textbf{91.40} & \textbf{98.50} & 92.62 & 97.15 & 99.18 & \textbf{87.59}\\
       LSTM & 77.17 & 81.96 & 93.73 & \underline{\textbf{93.68}} & \underline{\textbf{98.48}} & \textbf{99.36} & 85.42\\
       Transformer & 71.51 & 77.08 & 89.43 & 92.23 & 97.30 & 98.53 & 81.87\\
       \midrule
       \textbf{Translation} & & & & & & \\
       \midrule
       GCN & 85.89 & 94.73 & 98.85 & 90.10 & 92.45 & 95.61 & 87.99\\
       GraphSAGE & 88.15 & 96.31 & 99.25 & 90.68 & 94.51 & 96.88 & 89.41\\
       \midrule
       CNN & 84.72 & 94.66 & 98.70 & 90.17 & 93.98 & 97.96 & 87.44 \\
       LSTM& \underline{\textbf{89.85}} & \underline{\textbf{96.70}} & \underline{\textbf{99.35}} & 89.74 & 94.60 & 97.91 & \underline{\textbf{89.79}}\\
       Transformer& 86.64  & 95.78 &  98.83 &  \textbf{90.93}  &  \textbf{96.05} & \underline{\textbf{99.73}} & 88.78 \\
       \bottomrule
        \end{tabular}
    \caption{Overall performance of different neural encoders and methods for encoding multiple mathematical operations (i.e., integration, differentiation, addition, difference, multiplication, division) in the latent space.}
    \label{tab:compare_approaches_overall}
\end{table*}

\subsection{Operation Encoders}
The operation encoders are implemented using a lookup table similar to word embeddings \cite{mikolov2013efficient}, where each entry corresponds to the vector of a mathematical operator. We experiment with \emph{dense}\footnote{\url{https://pytorch.org/docs/stable/generated/torch.nn.Embedding.html}} embeddings for the translation model and instantiate the projection architecture with both \emph{dense} and \emph{one-hot}\footnote{\url{https://pytorch.org/docs/stable/generated/torch.nn.functional.one_hot.html}} embeddings. The translation model requires the operation embeddings to be the same size as the expression embeddings, admitting, therefore, only dense representations.

\section{Training Details}

As the models are trained to predict a target embedding, 
the main goal during optimisation is to avoid a \emph{representational collapse} in the expression encoder.
To this end, we opted for a \emph{Multiple Negatives Ranking (MNR)} loss with in-batch negative examples \cite{henderson2017efficient}. This technique allows us to sidestep the explicit selection of the negative sample, enabling a smoother optimisation of the latent space.   
We trained the models on a total of 12.800 premise expressions with 24 positive examples each derived from the application of 6 operations (see Section \ref{sec:synthetic_data}). This produces over 307.200 training instances composed of premise $x$, operation $t$, and conclusion $y$. The models are then trained for 32 epochs with a batch size of 64 (with in-batch random negatives). We found that the best results are obtained with a learning rate of 1e-5.

\section{Empirical Evaluation}

\subsection{Empirical Setup}
We evaluate the performance of different representational paradigms and expression encoders by building held-out dev and test sets. In particular, to assess the structural organisation of the latent space, we frame the task of multi-operational inference as an expression retrieval problem. 
Given a premise $x$, an operation $t$, a sample of positive conclusions $P = \{p_1, \ldots, p_n\}$, and a sample of negative conclusions $N = \{n_1, \ldots, n_m\}$, we adopt the models to predict an embedding $\textbf{e}_y'$ (Section \ref{sec:jepa}) and employ a distance function $\delta$ to rank all the conclusions in $P \cup N$ according to their similarity with $\textbf{e}_y'$. We implement $\delta$ using cosine similarity, and construct two evaluation sets to assess complementary inferential properties, namely:

\begin{figure*}[t]
\centering
\subfloat[Projection]{\includegraphics[width=0.45\textwidth]{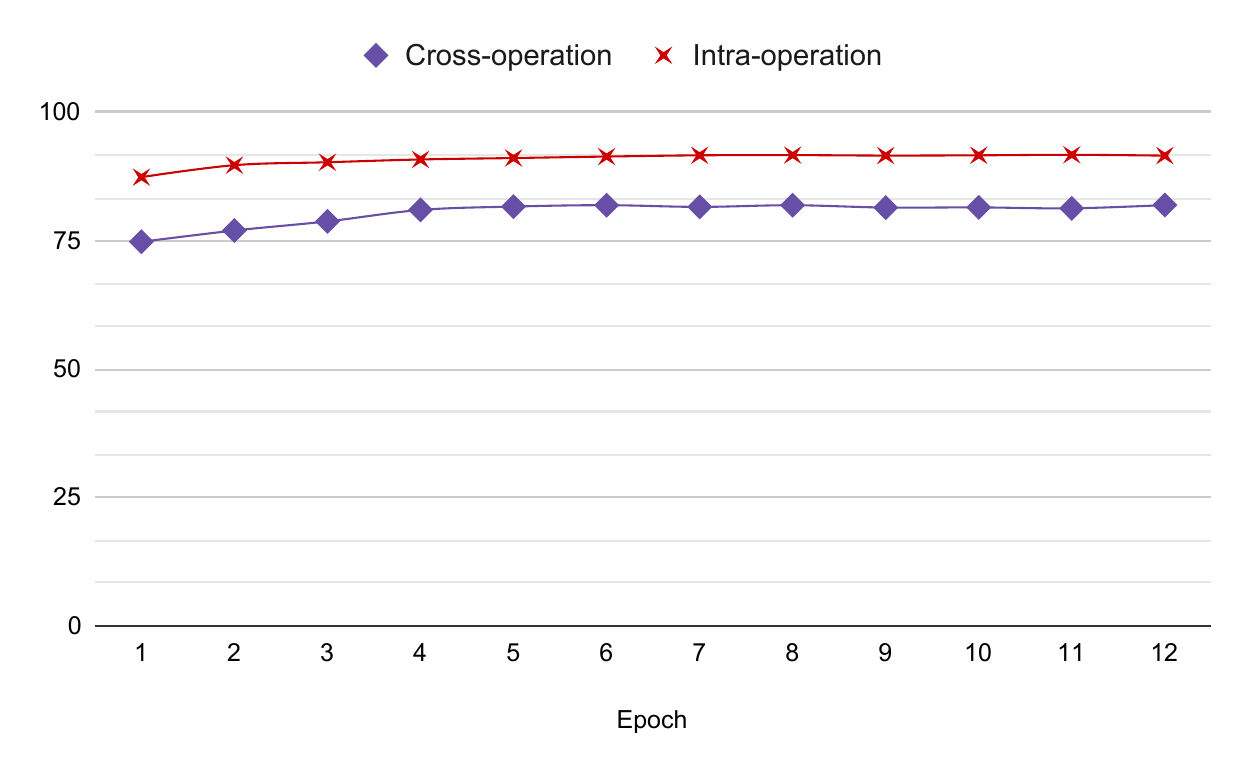}}
\subfloat[Translation]{\includegraphics[width=0.45\textwidth]{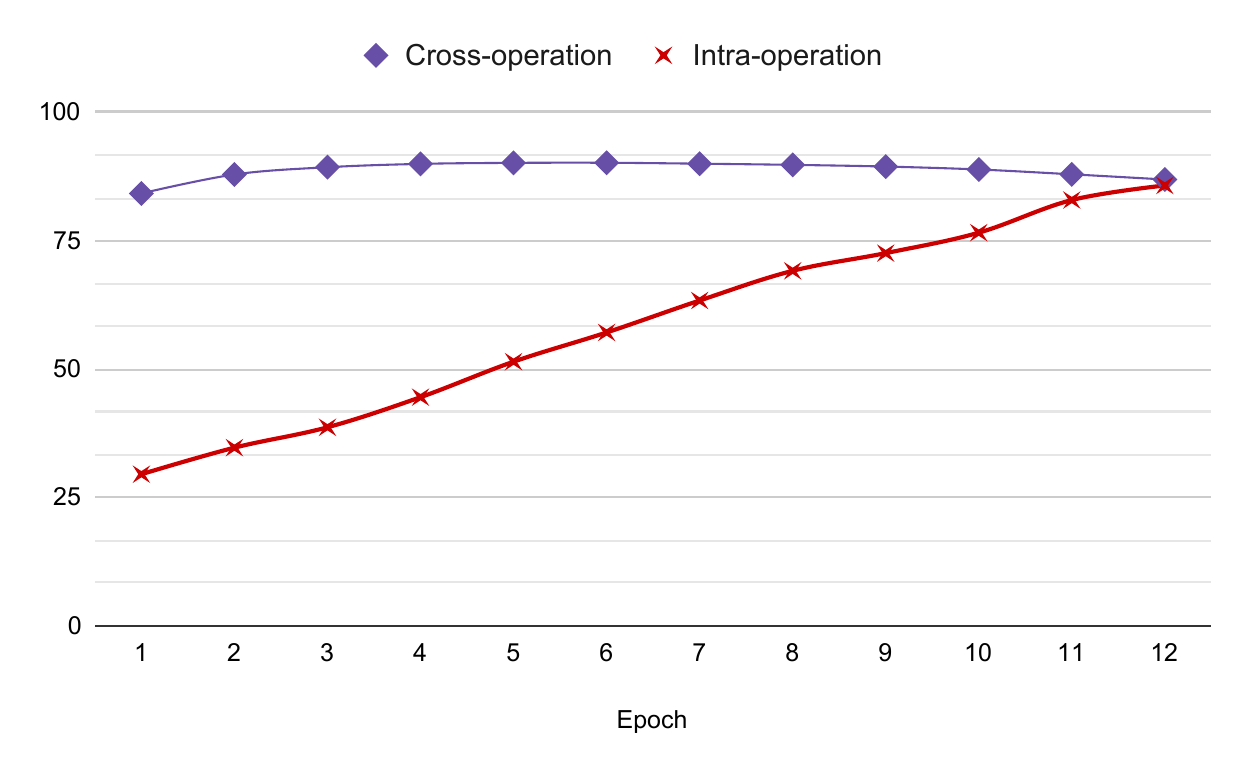}}
\caption{Typical training dynamics of different multi-operational paradigms (MAP on the dev set).}
\label{fig:training_dynamics}
\end{figure*}

\paragraph{Cross-operational Inference.} A model able to perform multi-operational inference should discriminate between the results of different operations applied to the same premise. Therefore, given a premise $x$ and an operation $t$ (e.g., addition), we construct the negative set $N$ by selecting the positive conclusions resulting from the application of different operations (e.g., differentiation, subtraction) to the same premise $x$. This set includes a total of 4 positive and 20 negative examples (extracted from the remaining 5 operations) for each premise-operation pair (for a total of 3k dev and 6k test instances).

\paragraph{Intra-operational Inference.} While we want the models to discriminate between different operators, a well-optimised latent space should still preserve the ability to predict the results of a single operation applied to different premises. Therefore, given a premise $x$ and an operation $t$, we construct the negative set $N$ by selecting the positive conclusions resulting from the application of the same operation $t$ to a different sample of premises. This set includes a total of 4 positive and 20 negative examples (extracted from 5 random premises) for each premise-operation pair (for a total of 3k dev and 6k test instances).

\paragraph{Metrics.}
The models are evaluated using Mean Average Precision (MAP) and Hit@k. Hit@k measures the percentage of test instances in which at least one positive conclusion is ranked within the top k positions. MAP, on the other hand, measures the overall ranking. We use the average MAP between cross-operational and intra-operational sets (dev) as a criterion for model selection.

\begin{figure*}[t]
    \centering
    \includegraphics[width=0.9\textwidth]{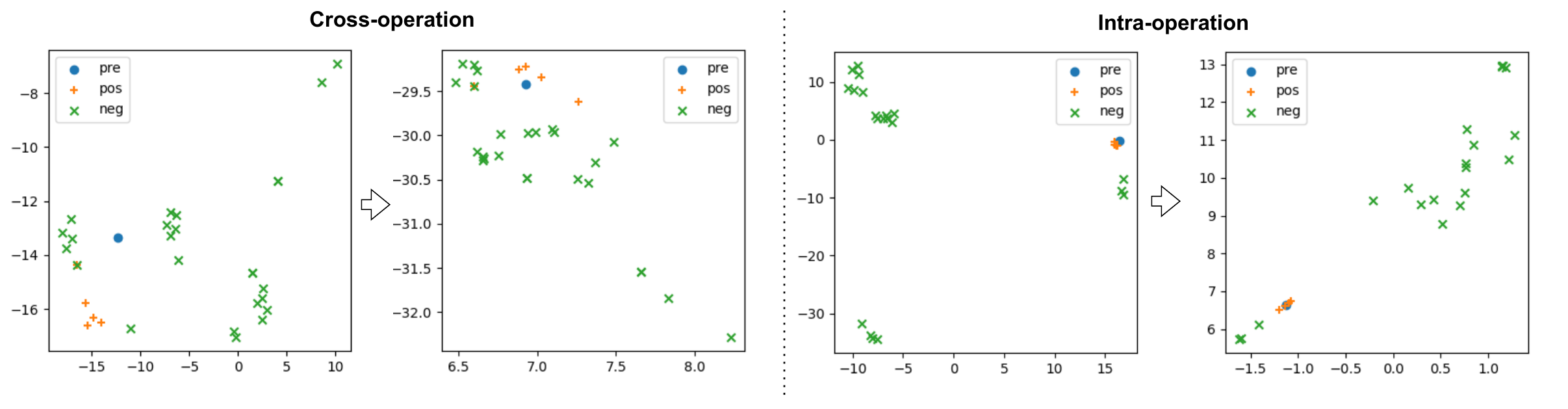}
    \caption{
    2D projection of the latent space before and after an operation-specific transformation. The visualization supports the crucial role of the multi-operational paradigm for \emph{cross-operational} inference, showing, at the same time, that \emph{intra-operational} inference concerns larger regions and can be achieved in the original expression encoder.
    }
\label{fig:visualization}
\end{figure*}

\subsection{Results}
Table \ref{tab:compare_approaches_overall} shows the performance of different encoders and paradigms on the test sets (i.e., evaluating the best models from the dev set, see Table \ref{tab:compare_approaches_overall_dev}). We can derive the following conclusions:

\paragraph{The translation mechanism improves cross-operational inference.} The models that use the translation method consistently outperform the models that use the projection method on the cross-operational inference task. This indicates that the translation paradigm can better capture the semantic relations between different operations and preserve them in the latent space. This is attested by the significant improvement achieved by different encoders, involving both graph-based and sequential architectures (e.g., +15.13\% and +7.64\%  for Transformers and GCN respectively).

\paragraph{Trade-off between cross-operational and intra-operational inference.} The models that excel at cross-operational inference tend to achieve lower performance on the intra-operational set. This suggests that there is a tension between generalising across different operations and specialising within each operation. Moreover, the results suggest that intra-operational inference represents an easier problem for neural encoders that can be achieved already with sparse multi-operational methods (i.e., models using one-hot projection can achieve a MAP score above 90\%).

\paragraph{LSTMs and GraphSAGE achieve the best performance.} LSTMs achieve the highest average MAP score, followed by GraphSAGE. These results demonstrate that LSTMs and GraphSAGE can balance between generalisation and specialisation, and leverage both sequential and graph-based information to encode mathematical operations. Moreover, we observe that graph-based models and CNNs tend to exhibit more stable performance across different representational paradigms (e.g., GraphSage achieve an average improvement of 2.3\%), while LSTMs and Transformers achieve balanced results only with the translation mechanism (i.e., with an average improvement of 4.37\% and 6.91\% respectively).

\paragraph{Model size alone does not explain inference performances.} The Transformer model, which has the largest number of parameters, exhibits a lower average MAP score (with the projection mechanism in particular). This implies that simply increasing the model complexity or capacity does not guarantee better results (see Table \ref{tab:compare_approaches_overall_dev} for additional details) and may compromise operational control in the latent space. This suggests that model architecture and the encoding method are more important factors for learning effective representations supporting multiple mathematical operations.

\subsection{Training Dynamics}

We conduct an additional analysis to investigate the training dynamics of different architectures. The graphs in Figure \ref{fig:training_dynamics} show the typical trend for the MAP achieved at different epochs on different evaluation sets. Interestingly, we found that \textbf{the projection and translation mechanisms optimise the latent space in a different way}. 
The projection paradigm, in fact, prioritises performance on intra-operational inference, with a constant gap between the two sets. Conversely, the translation paradigm supports a rapid optimisation of cross-operational inference, followed by a more gradual improvement on the intra-operational set. 

\begin{table}[t]
    \tiny
    \centering
    \begin{tabular}{l|cc|cc}
       \toprule
        & \textbf{Cross} & & \textbf{Intra} &\\
       \midrule
       \textbf{Proj. (1-hot)} & $\delta(\textbf{e}_x, \textbf{e}_y)$ & $\delta(\textbf{e}_{y}', \textbf{e}_y)$& $\delta(\textbf{e}_x, \textbf{e}_y)$ & $\delta(\textbf{e}_{y}', \textbf{e}_y)$\\
       \midrule
       GCN & 00.00 & 20.87  & 84.46 & 84.82\\
       GraphSAGE & 00.00 & 25.71 & 84.00 & 85.77\\
       \midrule
       CNN & 00.00 & 19.64 & 85.81 & 85.86\\
       LSTM & 00.00 & 24.11 & 86.12 & 84.87 \\
       Transformer & 00.00 & 16.52 & \textbf{88.20} & \textbf{86.80}\\
       \midrule
       \textbf{Proj. (Dense)} & $\delta(\textbf{e}_x, \textbf{e}_y)$ & $\delta(\textbf{e}_{y}', \textbf{e}_y)$& $\delta(\textbf{e}_x, \textbf{e}_y)$ & $\delta(\textbf{e}_{y}', \textbf{e}_y)$\\
       \midrule
       GCN & 00.00 & 22.88  & 84.38 & 85.25\\
       GraphSAGE & 00.00 & 25.47 & 84.96 & 86.01\\
       \midrule
       CNN & 00.00 & 23.13 & 83.38 & 82.84\\
       LSTM & 00.00 & 25.66 & 84.80 & 83.44 \\
       Transformer & 00.00 & 21.54 & 86.22 & 83.80\\
       \midrule
       \textbf{Translation} & $\delta(\textbf{e}_x, \textbf{e}_y)$ & $\delta(\textbf{e}_{y}', \textbf{e}_y + \textbf{t})$& $\delta(\textbf{e}_x, \textbf{e}_y)$ & $\delta(\textbf{e}_{y}', \textbf{e}_y + \textbf{t})$\\
       \midrule
       GCN & 00.00 & 11.85 & -07.13 & 39.76\\
       GraphSAGE & 00.00 & 11.37 & -01.45 & 40.98\\
       \midrule
       CNN & 00.00 & 05.23 & 12.32 & 33.68\\
       LSTM  & 00.00 & 40.20 & -00.46 & 51.46\\
       Transformer & 00.00 & \textbf{43.38} & 03.07 & 69.14\\
       \bottomrule
    \end{tabular}
    \caption{Latent separation of positive and negative examples before (i.e., $\delta(\textbf{e}_x, \textbf{e}_y)$) and after (i.e., $\delta(\textbf{e}_{y}', \textbf{e}_y)$) applying an \emph{operation-specific transformation}.}
    \label{tab:compare_latent}
\end{table}

This behaviour can help explain the difference in performances between the models. Specifically, since cross-operational inference is about disentangling operations applied to the same premise, we hypothesise it to require a more fine-grained optimisation in localised regions of the latent space. This optimisation can be compromised when priority is given to the discrimination of different premises, which, as in the case of intra-operational inference, involves a more coarse-grained optimisation in larger regions of the space.

\subsection{Latent Space Analysis}

\begin{figure*}[t]
\centering
\subfloat[Projection]{\includegraphics[width=0.45\textwidth]{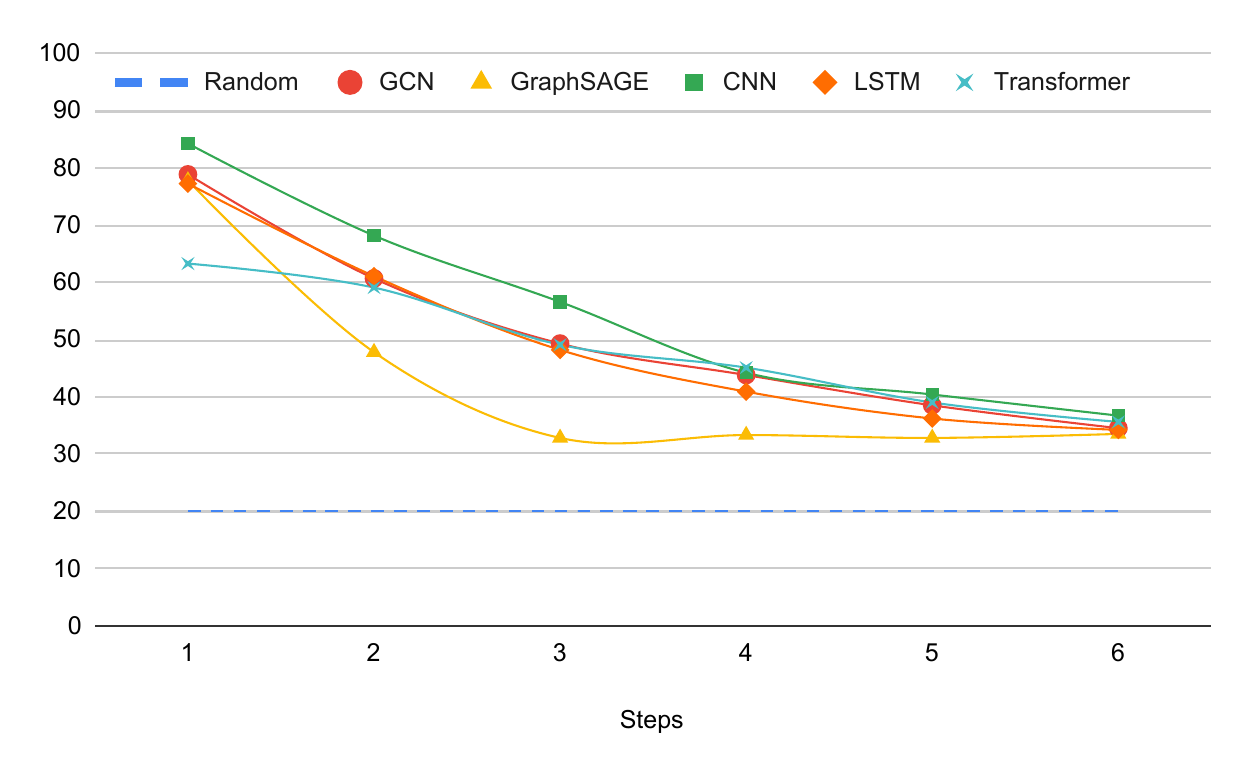}}
\subfloat[Translation]{\includegraphics[width=0.45\textwidth]{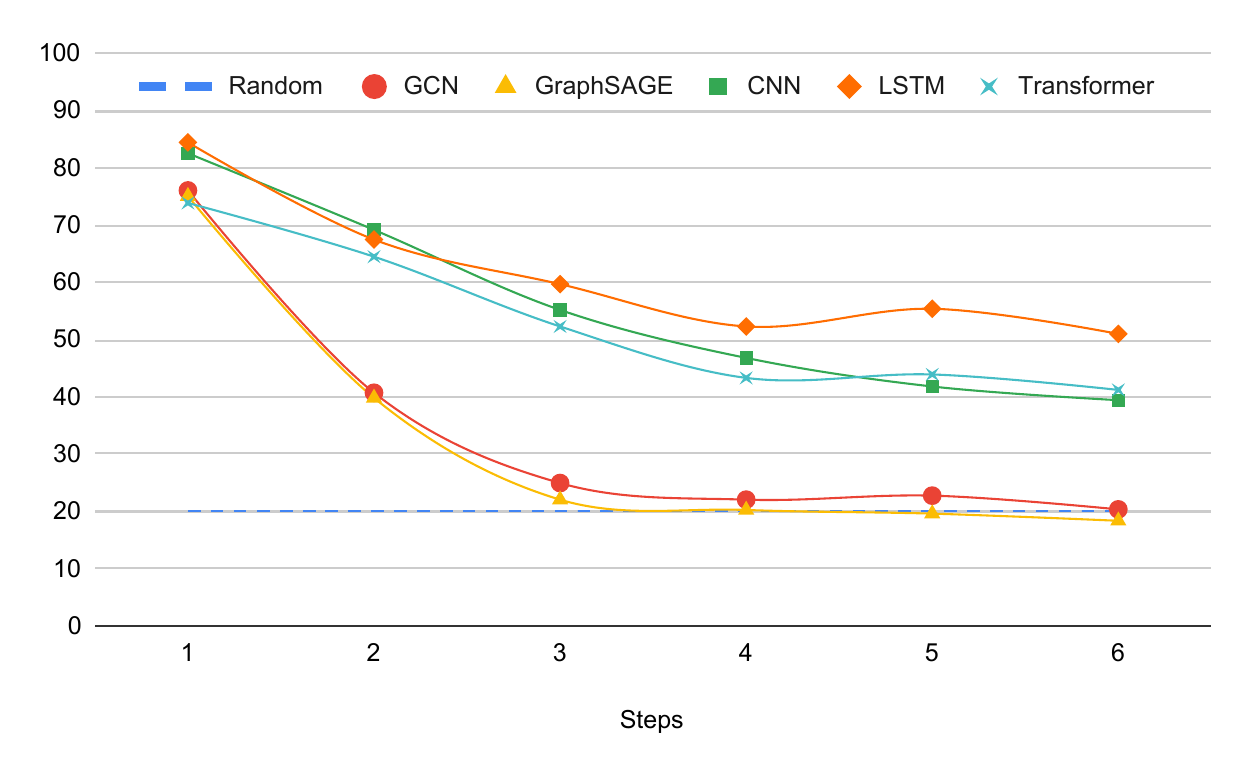}}
\caption{Multi-step derivations in latent space with different multi-operational paradigms and neural encoders.}
\label{fig:multistep}
\end{figure*}

\begin{figure*}[t]
\centering
\subfloat[Cross-operational]{\includegraphics[width=0.45\textwidth]{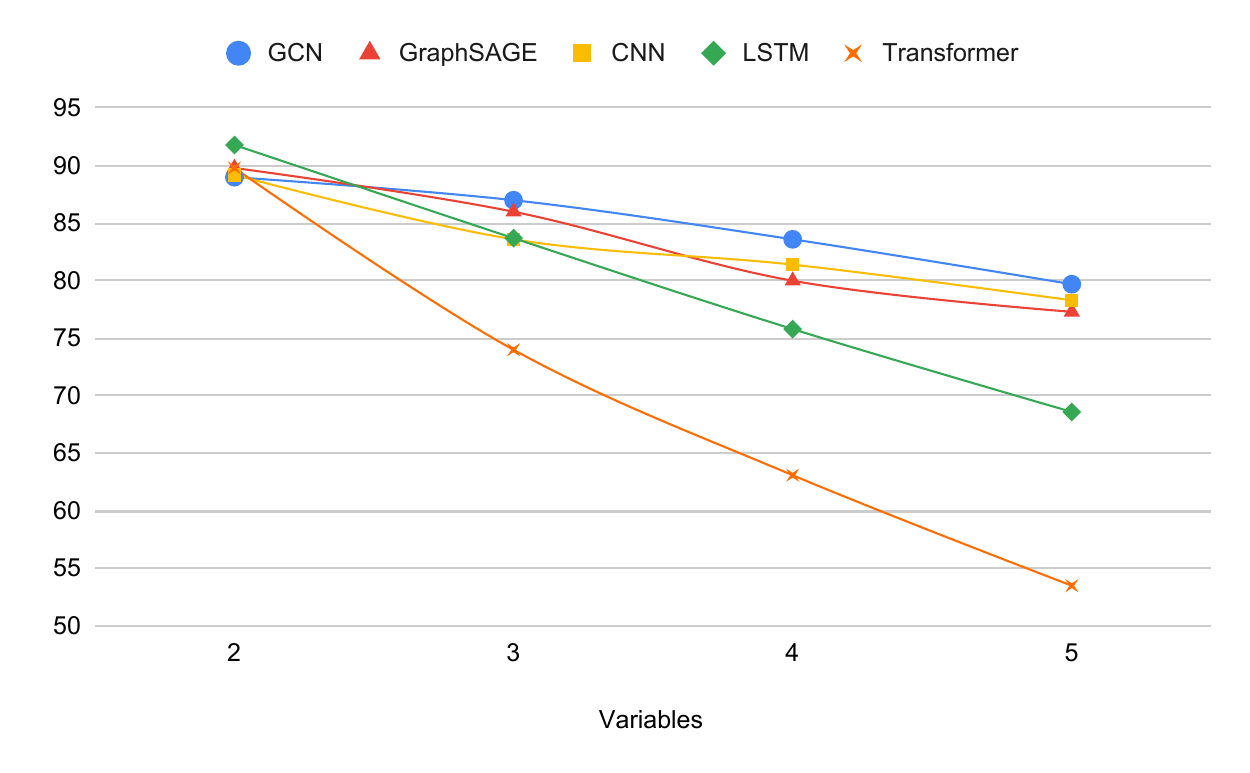}}
\subfloat[Intra-operational]{\includegraphics[width=0.45\textwidth]{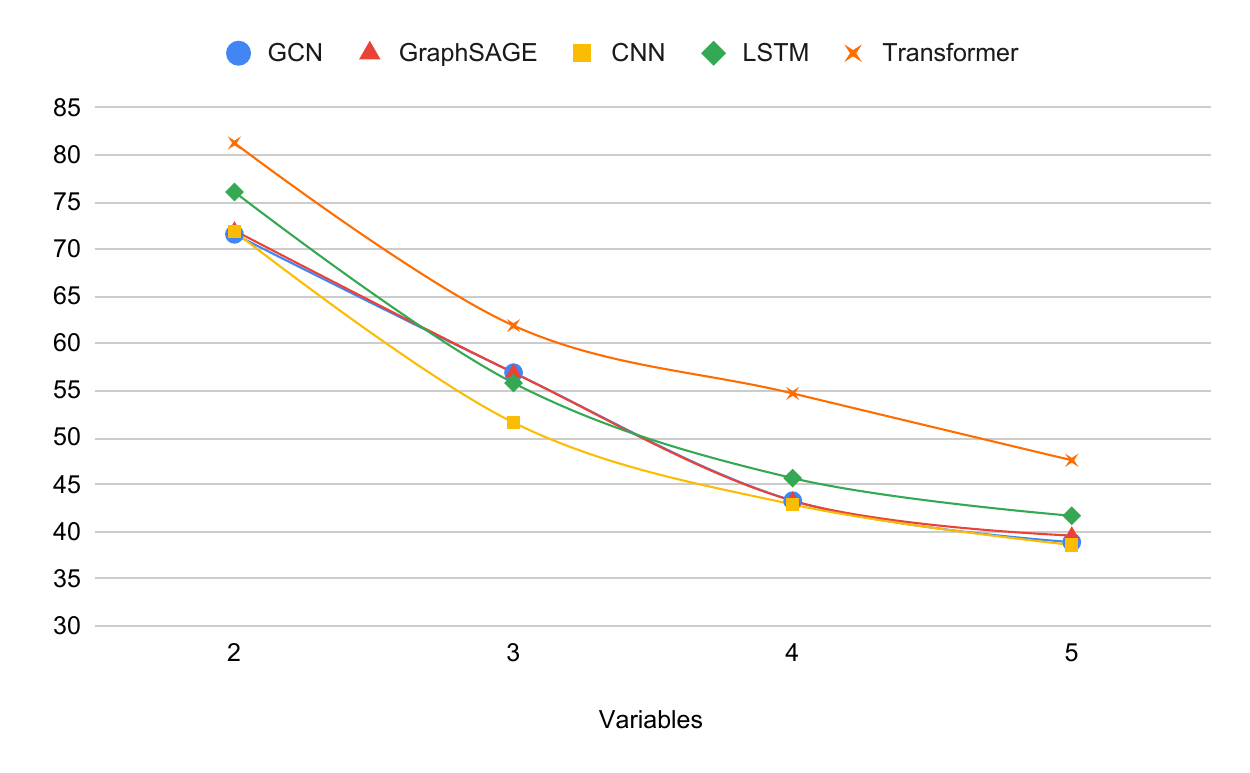}}
\caption{Length generalisation experiments by training different encoders (i.e., with translation) on premises with 2 variables, and testing on longer premises (MAP score).}
\label{fig:len_generalisation}
\end{figure*}

We further investigate this behaviour by measuring and visualising the latent space in the original expression encoder (i.e., computing $\delta(\textbf{e}_x, \textbf{e}_y)$) and after applying a transformation via the operation encoder (i.e., computing $\delta(\textbf{e}_y', \textbf{e}_y)$). In particular, Table \ref{tab:compare_latent} reports the average difference between the cosine similarity of the premises with positive and negative examples, a measure to estimate the latent space separation, and therefore, assess how dense the resulting vector space is.

From the results, we can derive the following main observations: 
\textbf{(1) The separation tends to be significantly lower in the cross-operational set}, confirming that the latent space requires a more fine-grained optimisation in localised regions (Fig. \ref{fig:visualization}); \textbf{(2) Cross-operational inference is not achievable without operation-specific transformations}, as confirmed by the impossibility to discriminate between positive and negative examples in the original expression encoders (i.e., $\delta(\textbf{e}_x, \textbf{e}_y)$, Table \ref{tab:compare_latent}); \textbf{(3) The projection mechanism achieves intra-operational separation in the original expression encoders.} This is not true for the translation mechanism in which the transformation induced by the operation encoder is fundamental for the separation to appear; \textbf{(4) The latent space resulting from the translation model is more dense}, with values for the separation that are generally lower when compared to the projection mechanism.

These results, combined with the performance in Table \ref{tab:compare_approaches_overall}, confirm that the translation paradigm can result in a more fine-grained and smoother optimisation which supports performance on cross-operational inference and a more balanced integration between expression and operation encoders. 

\subsection{Multi-Step Inference}
\label{sec:multi_step_derivations}

We investigate the behaviour of different encoders and representational paradigms when propagating latent operations for multiple steps. 
To experiment, we employ the architectures recursively by interpreting the predicted target embedding $\textbf{e}_y'$ as a premise representation for the next step (see Fig. \ref{fig:architectures}). In this case, we evaluate the performance using Hit@1, selecting 1 positive example and 4 negative examples for each premise and derivation step (2 for cross-operational and 2 for intra-operational). 

Figure \ref{fig:multistep} shows the obtained results. We found that \textbf{the majority of the models exhibit a latent organisation that allows for a non-random propagation of latent mathematical operations}. Most of the encoders, in fact, achieve performances that are significantly above random performance after 6 latent derivation steps (with a peak of 30\% improvement for LSTM + translation). Moreover, while all the models tend to decrease in performance with an increasing number of inference steps, \textbf{we observe significant differences between paradigms and classes of encoders}. Most notably, we found that \textbf{the performance of graph-based encoders tends to decrease faster, while the sequential models can obtain more stable results, in particular with the translation paradigm}. The best translation model (i.e., LSTM) achieves a Hit@1 score at 6 steps of up to 50\%, that is $\approx$15\% above the best projection architecture (i.e., CNN).

\subsection{Length Generalisation}

Finally, we perform experiments to test the ability of expression encoders to generalise to out-of-distribution examples. In particular, we focus on length generalisation which constitutes a notoriously hard problem for neural networks \cite{shen2021towards,hupkes2020compositionality,geirhos2020shortcut}. To this end, we train the models on the subset of the training set containing premises with 2 variables and assess performance on longer premises (i.e., grouping the test set according to the number of variables). Figure \ref{fig:len_generalisation} shows the results for different encoders using the translation mechanism.

Overall, the results show \textbf{a decrease in performance as expected, demonstrating, at the same time, a notable difference between encoders on cross-operational inference}. In particular, the results suggest that \textbf{graph-based models can generalise significantly better on longer premises}, probably due to their ability to capture explicit hierarchical dependencies within the expressions. \textbf{Among the sequential models, CNNs achieve better generalisation performance}. We attribute these results to the convolution operation in CNNs which may help capture structural invariances within the expressions and allow a generalisation that is similar to GCNs.

\subsection{Discussion}

From the empirical evaluation, we can derive a set of takeaways for both the joint-embedding architectures and the specific expression encoders.

Regarding the architectures, our analysis suggests that the translational paradigm can result in a more fine-grained and smoother optimisation of the latent space (Figure \ref{fig:training_dynamics} and Table \ref{tab:compare_latent}). This has the effect of improving multi-operational inference enabling a more balanced integration of different expression encoders, with an overall better trade-off between cross-operational and intra-operational inference (Table \ref{tab:compare_approaches_overall}). Moreover, we found that the translational paradigm can support better generalisation on multi-step inference when instantiated with sequential encoders such as Transformers, CNNs, and LSTMs (Figure \ref{fig:multistep}), even when the encoders are only trained on single-step derivations.

Regarding the specific encoders, we conclude that different models have different characteristics that should inform practitioners and future research in the field. Sequential models (i.e., Transformers, CNNs, and LSTMs), possess a better ability to organise the latent space for enabling latent multi-step derivations (Figure \ref{fig:multistep}). Conversely, graph-based models are more efficient (i.e., they achieve better performance using smaller operation encoders, see one-hot in Table \ref{tab:compare_approaches_overall}) and tend to generalise better to longer expressions when trained to simpler ones (see Figure \ref{fig:len_generalisation}).

\section{Related Work}

The quest to understand whether neural architectures can perform mathematical reasoning has led researchers to investigate several tasks and evaluation methods \cite{lu-etal-2023-survey,meadows2023introduction,mishra-etal-2022-lila,ferreira-etal-2022-integer,ferreira2020premise,welleck2021naturalproofs,valentino-etal-2022-textgraphs,mishra2022numglue,petersen-etal-2023-neural}.
In this work, we focused on equational reasoning, a particular instance of mathematical reasoning involving the manipulation of expressions through the systematic application of specialised operations \cite{welleck2022symbolic,lample2019deep,saxton2018analysing}.
In particular, our work is inspired by previous attempts to approximate mathematical reasoning entirely in latent space \cite{lee2019mathematical}. Differently from \citet{lee2019mathematical}, we investigate the joint approximation of multiple mathematical operations for expression derivation (\citet{lee2019mathematical} explore exclusively the rewriting operation for theorem proving). Moreover, while \citet{lee2019mathematical} focus on the evaluation of Graph Neural Networks \cite{paliwal2020graph}), we analyse the behaviour of a diverse set of representational paradigms and neural encoders.
Our data generation methodology is inspired by recent work leveraging symbolic engines and algorithms to build systematic benchmarks for neural models \cite{meadows2023symbolic,meadows2023generating,chen2022program,saparov2023testing}. However, to the best of our knowledge, we are the first to construct and release a synthetic dataset to investigate multi-step and multi-operational derivations in latent space.

\section{Conclusion}
This paper focused on equational reasoning for expression derivation to investigate the possibility of approximating and composing multiple mathematical operations in a single latent space. Specifically, we investigated different representational paradigms and encoding mechanisms, analysing the trade-off between encoding different mathematical operators and specialising within single operations, as well as the ability to support multi-step derivations and out-of-distribution generalisation.
Moreover, we constructed and released a large-scale dataset comprising 1.7M derivation steps stemming from 61K premises and 6 operators, which we hope will encourage researchers to explore future work in the field.

\section{Limitations}

The systematic application of mathematical operators requires reasoning at an intentional level, that is, the execution and composition of mathematical functions defined on a potentially infinite set of elements. Neural networks, on the contrary, operate at an extensional level and, by their current nature, can only approximate such functions by learning from a finite set of examples. 

Due to this characteristic, this work explored architectures that are trained on expressions composed of a predefined number and set of variables (i.e., between 2 and 5) and operators (i.e., addition, subtraction, multiplication, division, integration, differentiation), and, therefore, capable of performing approximation over a finite vocabulary of symbols. 
Extending the architectures with a new set of operations and out-of-vocabulary symbols, therefore, would require re-training the models from scratch. Future work could investigate this limitation by exploring, for instance, transfer learning techniques and more flexible neural architectures. 

For the same reason, we restricted our investigation to the encoding of atomic operations, that is, operations in which the second operand is represented by a variable. While this limitation is circumvented by the sequential application of operators in a multi-step fashion, this work did not explore the encoding of single-step operations involving more complex operands (e.g., multiplication between two expressions composed of multiple variables each). In principle, however, the evaluation presented in this work can be extended with the new synthetic data to accommodate and study different cases and setups in the future.  

\section*{Acknowledgements}
This work was partially funded by the Swiss National Science Foundation (SNSF) project NeuMath (\href{https://data.snf.ch/grants/grant/204617}{200021\_204617}), by the EPSRC grant EP/T026995/1 entitled “EnnCore: End-to-End Conceptual Guarding of Neural Architectures” under Security for all in an AI enabled society, by the CRUK National Biomarker Centre, and supported by the Manchester Experimental Cancer Medicine Centre.

\bibliography{anthology,custom}

\appendix

\section{Data Generation}
\label{sec:details_generation}

Algorithm 1 formalises the general data generation methodology adopted for generating premises with the SymPy\footnote{\url{https://www.sympy.org/en/index.html}} engine. 

\begin{algorithm}[t]
\caption{Premise Generation}
\begin{algorithmic}[1]
\State $\mathcal{F} \gets \text{premise.free\_symbols}$
\State $p \gets \frac{1}{p_r} - 1$
\For{$s$ in $\mathcal{F}$}
    \State $m \gets \text{random.choice}([0]*p + [1])$
    \If{$m = 0$}
        \State $s' \gets s$
    \Else
        \State $p_c \gets \frac{1}{p_e} - 1$
        \State $m \gets \text{random.choice}([0]*p_c + [1])$
        \State $c \gets \text{random.choice}(\{2, ..., 9\})$
        \If{$m = 0$}
            \If{\text{random.choice}(\{0,1\}) = 0}
                \State $s' \gets s \times c$
            \Else
                \State $s' \gets \frac{s}{c}$
            \EndIf
        \Else
            \State $c \gets \text{random.choice}(\{2, ..., 9\})$
            \State $s' \gets s^c$
        \EndIf
    \EndIf
    \State $\text{premise} \gets \text{premise.subs}(s, s')$
\EndFor
\State \Return premise
\end{algorithmic}
\label{alg:randomisation}
\end{algorithm}

The following is an example of an entry in the dataset with both LaTex and Sympy surface form for representing expressions, considering \emph{integration} and a single variable operand $r$. The same overall structure is adopted for the remaining operations and a larger vocabulary of variables:

\begin{itemize}
    \item Premise: 
    \begin{itemize}
            \item Latex: 
            $$u + \\cos{(\\log{(- x + o)})}$$
            \item SymPy: 
            $$Add(Symbol('u'),$$
           $$ cos(log(Add(Mul(Integer(-1),
           $$
           $$Symbol('x')),
           $$
           $$
           Symbol('o'))))) 
            $$
    \end{itemize}
    \item Derivation (\emph{integration}, $r$):
    \begin{itemize}  
            \item Latex: 
            $$u r + r \\cos{(\\log{(- x + o)})}$$
            \item SymPy: 
            $$ 
            Add(Mul(Symbol('u'), Symbol('r')), 
            $$
            $$
            Mul(Symbol('r'), cos(log(
            $$
            $$
            Add(Mul(Integer(-1), 
            $$
            $$
            Symbol('x')), Symbol('o'))))))
            $$
    \end{itemize}
\end{itemize}

\section{Dev Results}

\begin{table*}[th]
    \tiny
    \centering
    \begin{tabular}{c|ccc|ccc|c|cc}
        \toprule
         \textbf{}  & \textbf{MAP} & \textbf{Hit@1} & \textbf{Hit@3} & \textbf{MAP} & \textbf{Hit@1} & \textbf{Hit@3} & \textbf{Avg MAP} & \textbf{Embeddings Dim.} & \textbf{Model Size (MB)}\\ 
       \midrule
       \textbf{Projection (One-hot)} & & \textbf{Cross-op.} & & & \textbf{Intra-op.} & & &\\
       \midrule
       GCN & 71.37 & 74.86 & 86.46 & 92.80 & 97.83 & 99.03 & 82.20 & 300 & 3.0\\
        & 73.44 & \textbf{78.73} & 89.50 & 92.72 & 97.70 & \underline{\textbf{99.86}} & 83.08 & 512 & 7.9\\
       & 74.12 & 78.40 & 89.66 & 92.68 & 97.46 & 99.13 & 83.40 & 768 & 18.0\\
       GraphSAGE & 79.54 & 82.60 & 92.83 & 91.98 & 96.43 & 98.63 & 85.76 & 300 & 5.0\\
       & 81.57 & 84.90 & 95.00 & 92.81 & 97.50 & 99.20 & 87.19 & 512 & 14.0\\
       & \textbf{83.70} & \textbf{88.56} & \textbf{96.73} & \textbf{93.00} & 97.70 & 99.30 & \textbf{88.35} & 768 & 31.0\\
       \midrule
       CNN& 69.84 & 77.53 & 95.00 & 91.79 & 96.10 & 98.30 &  80.81 & 300 & 2.5\\
       & 66.94 & 74.46 & 93.40 & 92.06 & 96.80 & 98.40 & 79.50 & 512 & 4.6\\
       & 67.25 & 74.70 & 93.63 & 91.48 & 95.46 & 97.73 & 79.37 & 768 & 7.6\\
       LSTM & 69.31 & 72.06 & 89.00 & 92.93 & \textbf{97.96} & 99.46 & 81.12 & 300 & 6.6\\
       & 69.33 & 71.66 & 88.16 & 92.92 & 97.90 & 99.46 & 81.13 & 512 & 19.0\\
       & 70.84 & 72.60 & 90.10 & 92.89 & 97.76 & 99.30 & 81.86 & 768 & 42.0\\
       Transformer& 48.61 & 48.20 & 63.70 & 91.79 & 96.60 & 99.43 & 70.20 & 300 & 38.0\\
       & 46.29 & 43.70 & 61.30 & 91.72 & 96.46 & 99.16 & 69.01 & 512 & 75.0\\
       & 46.17 & 43.56 & 62.43 & 91.98 & 96.90 & 99.20 & 69.08 & 768 & 130.0\\
       \midrule
       \textbf{Projection (Dense)} & & \textbf{Cross-op.} & & & \textbf{Intra-op.} & & & \\
        \midrule
       GCN & 77.37 & 82.16 & 93.63 & 91.28 & 96.43 & 98.93 & 84.33 & 300 & 3.3\\
       & 77.89 & 83.46 & 92.70 & 92.45 & 97.33 & 98.90 & 85.17 & 512 & 8.9\\
       & 79.93 & 85.63 & 94.10 & 92.09 & 96.93 & 99.00 & 86.01 & 768 & 20.0\\
       GraphSAGE  & 81.39 & 84.83 & 95.76 & 91.08 & 95.80 & 98.60 & 86.24 & 300 & 5.4\\
       & 80.73 & 84.06 & 94.20 & 92.36 & 97.10 & 98.86 & 86.54 & 512 & 15.0\\
       & 81.09 & 83.93 & 94.36 & 92.40 & 97.40 & 99.10 & 86.75 & 768 & 33.0\\
       \midrule
       CNN& 
       81.91 & 90.46 & 97.80 & 91.67 & 95.76 & 98.23 & 86.79 & 300 & 2.8\\
       & \textbf{82.70} & \textbf{92.10} & \textbf{98.73} & 91.89 & 95.93 & 98.33 & \textbf{87.30} & 512 & 5.6\\
       & 81.70 & 90.73 & 97.80 & 92.36 & 97.20 & 98.90 & 87.03 & 768 & 9.9\\
       LSTM & 71.96 & 74.93 & 89.03 & 92.26 & 96.93 & 99.26 & 82.11 & 300 & 7.0\\
       & 76.13 & 80.50 & 93.53 & 92.77 & 97.70 & 99.30 & 84.45 & 512 & 20.0\\
       & 76.40 & 80.03 & 93.23 & \underline{\textbf{93.13}} & \underline{\textbf{98.06}} & \textbf{99.60} & 84.76 & 768 & 44.0\\
       Transformer
       & 70.06 & 75.50 & 88.70 & 91.96 & 97.40 & 99.53 & 81.01 & 300 & 38.0\\
        &  69.59 &  73.63 & 87.20 &  92.20 & 97.70  &  99.53 & 80.89 & 512 & 76.0\\
       & 52.43 & 51.16 & 68.30 & 90.78 & 96.16 & 99.33 & 71.60 & 768 & 133.0\\
        \midrule
       \textbf{Translation} & & \textbf{Cross-op.} & & & \textbf{Intra-op.} & & & \\
       \midrule
       GCN & 80.16 & 89.50 & 96.63 & 83.90 & 86.83 & 94.16 & 82.03 & 300 & 3.0\\
       & 86.56 & 95.03 & 99.03 & 86.20 & 89.16 & 96.16 & 86.38 & 512 & 7.9\\
       & 86.72 & 95.53 & 99.30 & 87.85 & 91.40 & 96.50 & 87.29 & 768 & 18.0\\
        GraphSAGE & 84.94 & 92.93 & 98.10 & 84.13 & 86.00 & 93.40 & 84.53 & 300 & 5.0\\
       & 87.44 & 95.26 & 99.00 & 88.70 & 92.76 & 96.63 & 88.07 & 512 & 14.0\\
       & 88.39 & 96.13 & 99.16 & 90.35 & 94.20 & 97.86 & 89.37 & 768 & 31.0\\
        \midrule
       CNN& 84.42 & 95.30 & 98.93 & 89.22 & 93.66 & 97.66 & 86.81 & 300 & 2.5\\
       & 84.62 & 95.33 & 99.20 & 90.36 & 96.03 & 98.63 & 87.49 & 512 & 4.6\\
       & 86.99 & 96.76 & \underline{\textbf{99.60}} & 87.18 & 93.46 & 96.90 & 87.08 & 768 & 7.7\\
       LSTM & 84.20 & 92.23 & 99.10 & 87.24 & 90.76 & 96.93 & 85.72 & 300 & 6.6\\
       & 86.98 & 94.36 & 99.06 & 88.81 & 93.70 & 98.00 & 87.89 & 512 & 19.0\\
       & \underline{\textbf{89.50}} & \underline{\textbf{97.13}} & 99.36 & 89.89 & 95.70 & 98.53 & \underline{\textbf{89.70}} & 768 & 42.0\\
       Transformer& 84.17 & 94.90 & 98.83 & 90.35 & 96.23 & 99.40 & 87.26 & 300 & 38.0\\
       & 85.99  & 95.70 &  98.70 &  \textbf{91.86}  &  \textbf{97.83} & \textbf{99.60} &  88.93  & 512 & 75.0\\
       & 86.04 & 95.33 & 98.53 & 91.27 & 97.03 & 99.20 & 88.65 & 768 & 130.0\\
       \bottomrule
        \end{tabular}
    \caption{Overall performance of different neural encoders and methods (dev set) for jointly encoding multiple mathematical operations (i.e., integration, differentiation, addition, difference, multiplication, division).}
    \label{tab:compare_approaches_overall_dev}
\end{table*}

Table \ref{tab:compare_approaches_overall_dev} reports the complete results on the dev set for different models and architectures with different embedding sizes.

\end{document}